\documentclass[acmsmall]{acmart}
\usepackage[utf8]{inputenc}
\usepackage{array}
\usepackage{longtable}
\usepackage{graphicx}
\usepackage{makecell}
\usepackage{booktabs}
\usepackage{float}
\usepackage{xcolor}
\usepackage{pdflscape}  

\AtBeginDocument{%
  }

\setcopyright{acmlicensed}
\copyrightyear{2025}
\acmYear{2025}
\acmDOI{XXXXXXX.XXXXXXX}

\acmJournal{JACM}
\acmVolume{37}
\acmNumber{4}
\acmMonth{5}

\begin{document}

\title{A Review on Influx of Bio-Inspired Algorithms: Critique and Improvement Needs}


\author{Shriyank Somvanshi}
\affiliation{%
  \institution{Texas State University}
  \city{San Marcos}
  \country{USA}}
\email{jum6@txstate.edu}

\author{Md Monzurul Islam}
\affiliation{%
  \institution{Texas State University}
  \city{San Marcos}
  \country{USA}}
\email{monzurul@txstate.edu}

\author{Syed Aaqib Javed}
\affiliation{%
  \institution{Texas State University}
  \city{San Marcos}
  \country{USA}}
\email{aaqib.ce@txstate.edu}

\author{Gaurab Chhetri}
\affiliation{%
  \institution{Texas State University}
  \city{San Marcos}
  \country{USA}}
\email{gaurab@txstate.edu}

\author{Kazi Sifatul Islam}
\affiliation{%
  \institution{Texas State University}
  \city{San Marcos}
  \country{USA}}
\email{kazi_sifat@txstate.edu}

\author{Tausif Islam Chowdhury}
\affiliation{%
  \institution{Texas State University}
  \city{San Marcos}
  \country{USA}}
\email{tausif.islam@txstate.edu}

\author{Sazzad Bin Bashar Polock}
\affiliation{%
  \institution{Texas State University}
  \city{San Marcos}
  \country{USA}}
\email{pay28@txstate.edu}

\author{Anandi Dutta, Ph.D.}
\affiliation{%
  \institution{Texas State University}
  \city{San Marcos}
  \country{USA}}
\email{anandi.dutta@txstate.edu}

\author{Subasish Das, Ph.D.}
\affiliation{%
  \institution{Texas State University}
  \city{San Marcos}
  \country{USA}}
\email{subasish@txstate.edu}

\renewcommand{\shortauthors}{Somvanshi et al.}

\begin{abstract}

 Bio-inspired algorithms utilize natural processes such as evolution, swarm behavior, foraging, and plant growth to solve complex, nonlinear, high-dimensional optimization problems. However, a plethora of these algorithms require a more rigor review before making them applicable to the relevant fields. This survey categorizes these algorithms into eight groups: evolutionary, swarm intelligence, physics-inspired, ecosystem and plant-based, predator–prey, neural-inspired, human-inspired, and hybrid approaches, and reviews their principles, strengths, novelty, and critical limitations. We provide a critique on the novelty issues of many of these algorithms. We illustrate some of the suitable usage of the prominent algorithms in machine learning, engineering design, bioinformatics, and intelligent systems, and highlight recent advances in hybridization, parameter tuning, and adaptive strategies. Finally, we identify open challenges such as scalability, convergence, reliability, and interpretability to suggest directions for future research. This work aims to serve as a resource for both researchers and practitioners interested in understanding the current landscape and future directions of reliable and authentic advancement of bio-inspired algorithms.
\end{abstract}




\received{20 May 2025}

\maketitle

\section{Introduction}

In an era marked by rapid technological advancement, the complexity of real-world computational problems, such as optimization, classification, scheduling, and control, has grown significantly. These problems are often characterized by high dimensionality, nonlinearities, dynamic environments, and incomplete or noisy data. Traditional optimization methods, including linear programming, gradient-based search, and exhaustive enumeration, frequently struggle in such settings due to their reliance on gradient information, rigid formulation requirements, and susceptibility to local optima. Their limitations are particularly evident in large-scale combinatorial tasks or non-differentiable solution spaces, where adaptability and global exploration are critical \cite{game_bio-inspired_2020}.

\subsection{Rise of Bio-Inspired Algorithms}
\textit{Bio-Inspired Algorithms (BIAs)} have emerged as a compelling alternative for addressing these challenges. Defined as a class of metaheuristic methods inspired by biological and natural processes, BIAs emulate strategies from evolution, swarm behavior, foraging, and immune systems. These algorithms are inherently stochastic, population-based, and adaptive, enabling them to traverse vast and complex search spaces efficiently. Fan et al.~\cite{fan_review_2020} categorize BIAs into evolutionary-based, swarm intelligence-based, ecology-based, and multi-objective optimization methods, while Sureka et al.~\cite{sureka_nature_2020} emphasize their resilience and adaptability under resource constraints. Their ability to avoid premature convergence, maintain diversity, and adapt to dynamic environments makes them particularly attractive for real-world problems \cite{cotta_bioinspired_2017, molina_paradox_2025, darwish_bio-inspired_2018}.

\subsection{Established Foundations}
A subset of BIAs-most notably Genetic Algorithms (GA), Evolution Strategies (ES), Differential Evolution (DE), Particle Swarm Optimization (PSO), and Ant Colony Optimization (ACO)-have achieved the status of well-established, rigorously validated methods. These approaches are grounded in sound theoretical principles, have been extensively benchmarked, and remain widely applied across domains ranging from robotics and engineering design to computational biology and distributed systems \cite{chu_solution_2011, ni_bioinspired_2016, heberle_bio-inspired_2011, binitha_survey_2012, janga_reddy_computational_2012, zheng_survey_2013}. Simon’s analytical treatment of Biogeography-Based Optimization (BBO) also illustrates how some newer methods, when rigorously framed, can be meaningfully positioned alongside classical evolutionary computation \cite{simon2011analytical}. These algorithms exemplify how bio-inspired computation, when theoretically justified, provides robust and transferable optimization strategies. 
Our review acknowledges these well-founded contributions as cornerstones, establishing a baseline against which newer methods must be judged.

\subsection{Proliferation of Metaphor-Based Methods}
Alongside these solid foundations, however, the field has witnessed an exponential proliferation of algorithms whose novelty is primarily metaphorical. Algorithms such as Harmony Search, Black Hole Optimization, Intelligent Water Drops, Firefly Algorithm, Bat Algorithm, Grey Wolf Optimizer, Salp Swarm Optimization, Cuckoo Search, and Grasshopper Optimization have been shown to be either reformulations or simplifications of existing evolutionary or swarm-based methods \cite{weyland2010rigorous, piotrowski2014novel, camacho2019intelligent, camacho2022analysis, camacho2023exposing, harandi2024grasshopper}. Detailed analyses reveal that many of these approaches introduce no fundamentally new operators or search principles, and instead rely on metaphor-driven terminology that obscures strong similarities to classical methods. Critical voices in the community argue that this trend risks fragmenting the field and diluting scientific rigor \cite{sorensen2015metaheuristics, fister2016new, aranha2022metaphor}. Molina et al.~\cite{molina_comprehensive_2020} further show that over one-third of published bio-inspired solvers are in fact versions of classical algorithms, underscoring the prevalence of redundancy.

\subsection{Balancing Promise and Criticism}
Despite this criticism, it would be misleading to dismiss the entire BIA field. Some metaphor-inspired methods have stimulated hybrid approaches, parameter tuning strategies, or domain-specific adaptations that yield practical benefits. Moreover, recent methodological works stress the importance of rigorous experimentation, benchmarking, and real-world validation to separate robust contributions from superficial proposals \cite{osaba2021tutorial}. Yet, as highlighted in multiple reviews and critical analyses, the unchecked proliferation of weakly justified algorithms risks undermining the credibility of bio-inspired computation as a whole \cite{camacho2023exposing, aranha2022metaphor, sorensen2015metaheuristics}. This survey therefore, takes a dual stance-recognizing legitimate contributions while critically assessing those that fall short of scientific rigor.

\subsection{Scope of This Survey}
\textbf{This paper aims to provide a critical synthesis of BIAs, highlighting both their enduring strengths and their ongoing pitfalls.} Specifically, it distinguishes between well-established algorithms with strong theoretical and empirical grounding and metaphor-driven approaches whose novelty remains questionable. In doing so, it contributes to a growing body of literature that calls for methodological rigor, principled evaluation, and consolidation in the field of bio-inspired computation. 
By doing so, we seek not only to map the landscape of BIAs but also to chart a clearer path forward, emphasizing consolidation over proliferation.

\textbf{The remainder of this paper is structured as follows:} Section~\ref{sec:evolution} outlines the historical evolution of major BIAs; Section~\ref{sec:taxonomy} revisits taxonomy with a critical lens; Section~\ref{sec:working} explains algorithmic mechanisms and variants; Section~\ref{sec:applications} explores applications across diverse domains; Section~\ref{sec:benchmarking} discusses benchmarking and reproducibility; Section~\ref{sec:challenges} highlights key criticisms and methodological weaknesses; Section~\ref{sec:future} suggests pathways forward; and Section~\ref{sec:conclusion} concludes with final reflections.


\section{Historical Evolution of Bio-Inspired Algorithms}
\label{sec:evolution}
  
The history of BIAs reflects both enduring methodological breakthroughs and the controversial rise of metaphor-driven variants. Since their inception, BIAs have been motivated by the need for adaptive, population-based approaches to nonlinear, high-dimensional, and dynamic optimization problems. As summarized in Table~\ref{tab:milestones_bio_algorithms}, this trajectory combines rigorously validated contributions that continue to inform optimization practice with numerous proposals later criticized for offering little beyond metaphorical novelty.

\subsection{Chronological Emergence}

The origins of BIAs can be traced to Holland’s Genetic Algorithm (GA) in 1975 \cite{Holland1975}, which established the principle of stochastic population-based search guided by natural selection. This milestone, along with subsequent developments in Evolution Strategies (ES) and Differential Evolution (DE), grounded the field in methods supported by schema theory, Markov models, and runtime analyses. In the 1990s, Ant Colony Optimization (ACO) \cite{dorigo1996ant} and Particle Swarm Optimization (PSO) \cite{kennedy1995particle} pioneered swarm intelligence by formalizing collective behaviors such as pheromone reinforcement and flocking. These algorithms remain benchmarks in optimization, with robust theoretical and empirical validation \cite{goldberg1989genetic, clerc2002particle}.

In the 2000s, the field expanded to encompass a wave of biologically and ecologically inspired methods. Bacterial Foraging Optimization (BFO) \cite{passino2002biomimicry} and Artificial Bee Colony (ABC) \cite{karaboga2008performance, karaboga2009survey} introduced microbial chemotaxis and honeybee foraging, respectively, as search mechanisms. Around the same time, a rapid proliferation of algorithms appeared, including Cuckoo Search (CS) \cite{yang2009cuckoo}, the Bat Algorithm (BA) \cite{yang2010bat}, Grey Wolf Optimizer (GWO) \cite{mirjalili2014grey}, Whale Optimization Algorithm (WOA) \cite{mirjalili2016whale}, Dragonfly Algorithm (DA) \cite{Mirjalili2016}, and Salp Swarm Algorithm (SSA) \cite{Mirjalili2017}. These methods broadened the landscape of BIAs by introducing increasingly specific biological analogies.

However, later analyses revealed that many of these newer methods offered little true novelty, often repackaging existing operators from GA, PSO, or DE with metaphorical framing. For example, SSA was shown to be non–shift invariant and underperformed even random search in certain cases \cite{castelli2022salp}; CS was demonstrated to be functionally equivalent to differential evolution and evolutionary strategies \cite{camacho2023exposing}; and GOA has been critiqued as a reformulation of PSO rather than a distinct algorithm \cite{harandi2024grasshopper}. Similar concerns were raised about BA, Firefly, and GWO, whose operators closely overlap with existing paradigms while lacking theoretical justification \cite{camacho2023exposing, aranha2022metaphor}. These findings underline that the evolution of BIAs is not a linear trajectory of innovation but a mixed landscape of substantive advances and superficial variants.

More recently, a constructive trend has emerged through hybrid models that integrate validated strategies rather than proposing entirely new metaphors. By combining techniques such as PSO, ABC, and GWO, hybrid BIAs address high-dimensional feature selection and other domain-specific challenges with greater robustness \cite{pham2023bio}. This reflects a pragmatic shift in the field: meaningful innovation stems less from novel metaphors and more from systematic integration and empirical validation.

\subsection{Motivation for Evolution}

The drivers of BIA development illustrate the tension between genuine performance needs and metaphorical creativity. Foundational algorithms such as GA, ES, DE, PSO, and ACO were motivated by the challenge of exploring complex search spaces without gradient information, maintaining diversity to avoid premature convergence, and balancing global exploration with local refinement \cite{Holland1975, kennedy1995particle, goldberg1989genetic}. Despite their success, these methods revealed limitations, including sensitivity to parameter settings and reduced accuracy near optima. This spurred the development of refinements such as memetic algorithms \cite{moscato1989evolution}, adaptive ACO variants \cite{blum2005ant}, and hybrid approaches that integrated multiple strategies.

By contrast, many later algorithms emerged under the justification of overcoming stagnation or improving adaptability, yet critical reviews show that their improvements were either incremental or unsupported by rigorous benchmarking. SSA’s conceptual flaws, CS’s lack of distinctiveness from classical models, and formal demonstrations that Harmony Search is reducible to existing strategies \cite{weyland2010rigorous} highlight this problem. As Fister et al.~\cite{fister2016new} and Sörensen \cite{sorensen2015metaheuristics} note, the field became saturated with superficially distinct algorithms motivated more by metaphorical novelty than substantive contribution.

\subsection{Critical Perspective on Progress}

Taken together, the historical trajectory of BIAs illustrates two intertwined narratives. On one hand, validated algorithms such as GA, ES, DE, PSO, ACO, and ABC represent enduring contributions that continue to influence optimization practice. On the other hand, the proliferation of metaphor-driven approaches such as CS, BA, GWO, SSA, and GOA reflects a problematic trend of superficial diversification, where novelty was often claimed through rebranding rather than operator-level innovation \cite{camacho2023exposing, aranha2022metaphor}. As Osaba et al.~\cite{osaba2021tutorial} emphasize, the future of BIAs depends not on unchecked proliferation but on rigorous benchmarking, reproducibility, and meaningful integration with other paradigms.

\begin{figure}[htp]
    \centering
    \includegraphics[width=0.98\linewidth]{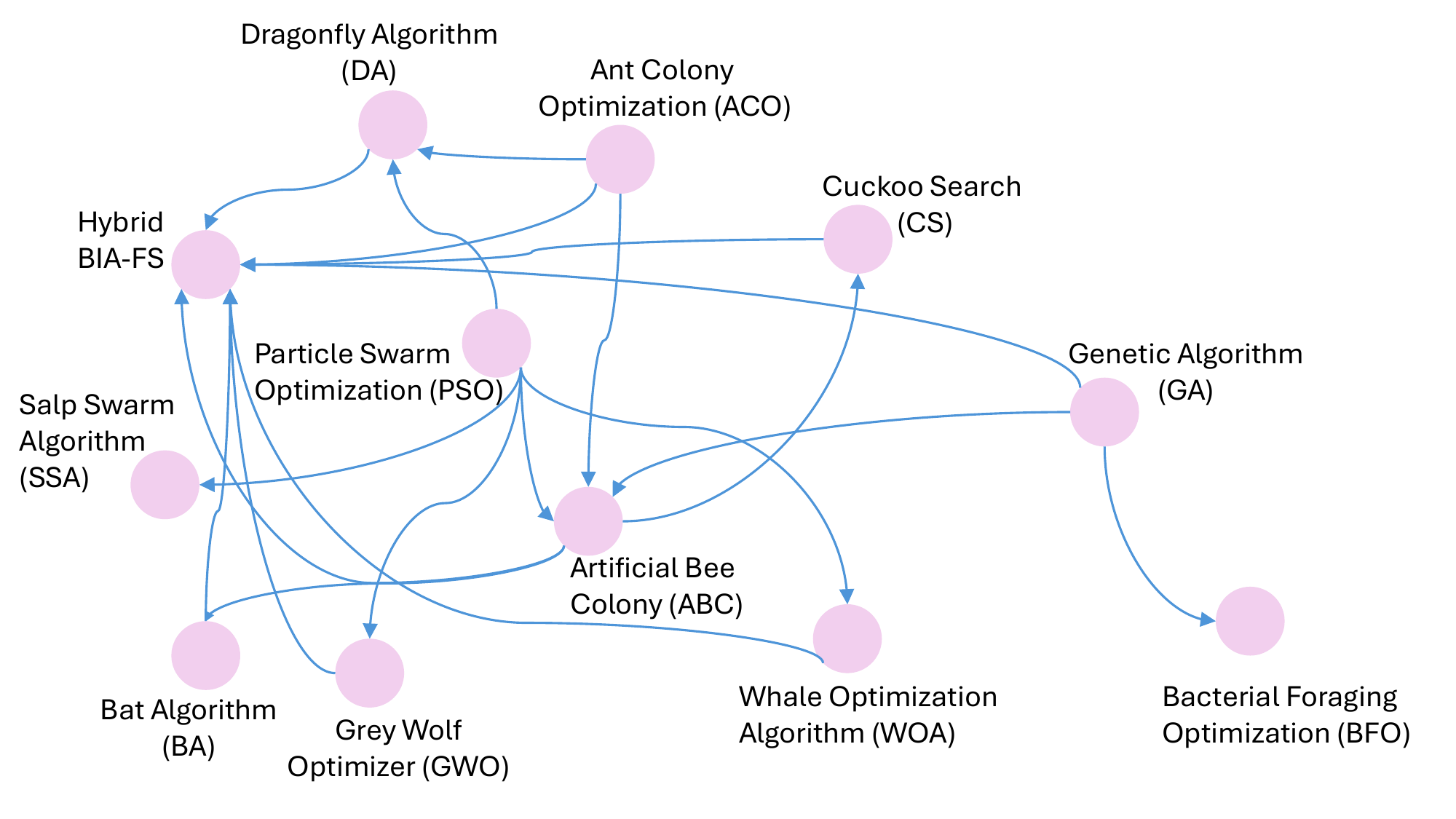}
    \caption{Evolutionary motivation and lineage of major bio-inspired algorithms}
    \label{fig:evolution_tree}
\end{figure}

Figure~\ref{fig:evolution_tree} reflects this dual perspective: foundational algorithms established the pillars of evolutionary computation and swarm intelligence, while later metaphor-inspired solvers often recycled existing principles under new biological analogies. Recognizing this divergence is critical for understanding both the genuine advances and the methodological pitfalls that have shaped the evolution of BIAs.

\vspace{-8pt}
\thispagestyle{empty}
\begingroup
\fontsize{8pt}{9pt}\selectfont
\begin{longtable}{m{0.08\linewidth} m{0.28\linewidth} m{0.18\linewidth} m{0.40\linewidth}}
\caption{Milestones in the evolution of Bio-Inspired Algorithms (BIAs), distinguishing well-validated foundations from contested metaphor-driven variants.} 
\label{tab:milestones_bio_algorithms} \\
\toprule
\textbf{Year} & 
\textbf{Algorithm} & 
\textbf{Status} & 
\textbf{Inspiration Source / Critical Notes} \\
\midrule
\endfirsthead
\toprule
\textbf{Year} & 
\textbf{Algorithm} & 
\textbf{Status} & 
\textbf{Inspiration Source / Critical Notes} \\
\midrule
\endhead
\midrule
\endfoot
\bottomrule
\endlastfoot

1975 & Genetic Algorithm (GA) \cite{Holland1975} & Well-validated foundation & Inspired by natural selection; supported by schema theory and Markov models. Remains a cornerstone of evolutionary computation. \\
\midrule
1992 & Ant Colony Optimization (ACO) \cite{dorigo1996ant} & Well-validated foundation & Based on ant foraging behavior; extensively benchmarked in routing, scheduling, and combinatorial optimization. \\
\midrule
1995 & Particle Swarm Optimization (PSO) \cite{kennedy1995particle} & Well-validated foundation & Inspired by bird flocking and fish schooling; widely applied and analyzed using stability and convergence frameworks. \\
\midrule
2002 & Bacterial Foraging Optimization (BFO) \cite{passino2002biomimicry} & Contested variant & Modeled on \textit{E. coli} chemotaxis; initially novel but later criticized for scalability issues and limited general impact. \\
\midrule
2005 & Artificial Bee Colony (ABC) \cite{karaboga2008performance, karaboga2009survey} & Well-validated foundation & Mimics honeybee foraging; accepted as a legitimate swarm model, though less theoretically mature than PSO/ACO. \\
\midrule
2009 & Cuckoo Search (CS) \cite{yang2009cuckoo} & Contested variant & Based on brood parasitism; later analyses showed equivalence to DE/ES with limited novelty \cite{camacho2023exposing}. \\
\midrule
2010 & Bat Algorithm (BA) \cite{yang2010bat} & Contested variant & Inspired by bat echolocation; critiqued as a stochastic reformulation of PSO/GA without unique operators \cite{camacho2023exposing}. \\
\midrule
2014 & Grey Wolf Optimizer (GWO) \cite{mirjalili2014grey} & Contested variant & Models wolf pack hunting hierarchy; overlaps heavily with PSO-like leadership structures, originality questioned. \\
\midrule
2016 & Whale Optimization Algorithm (WOA) \cite{mirjalili2016whale} & Contested variant & Simulates bubble-net feeding in whales; primarily metaphor-driven, lacks independent theoretical justification. \\
\midrule
2016 & Dragonfly Algorithm (DA) \cite{Mirjalili2016} & Contested variant & Captures swarming patterns (alignment, cohesion, separation); offers incremental differences over PSO, novelty debated. \\
\midrule
2017 & Salp Swarm Algorithm (SSA) \cite{Mirjalili2017} & Contested variant & Chain foraging of salps; shown to underperform and conceptually flawed relative to classical methods \cite{castelli2022salp}. \\
\midrule
2023 & Hybrid BIAs for Feature Selection \cite{pham2023bio} & Emerging constructive trend & Integrates validated algorithms (PSO, ABC, GWO) for feature selection; represents a shift toward pragmatic, domain-specific hybridization. \\
\end{longtable}
\endgroup

\section{Taxonomy and Categorization} 
\label{sec:taxonomy}

Taxonomy provides a systematic framework for organizing BIAs according to their sources of inspiration and operator design. Historically, taxonomies have been used to emphasize the diversity of metaphors in the field \cite{fan_review_2020, sureka_nature_2020}. However, as multiple critiques point out, the act of classification can also inadvertently legitimize methods that are metaphorically novel but technically redundant \cite{sorensen2015metaheuristics, fister2016new, camacho2023exposing, aranha2022metaphor}. In this section, we revisit BIA taxonomy not as a neutral catalog but as a diagnostic tool. We distinguish categories anchored by foundational contributions with lasting theoretical or empirical value from those populated primarily by contested variants.  

\begin{figure}[htp]
    \centering
    \includegraphics[width=0.98\linewidth]{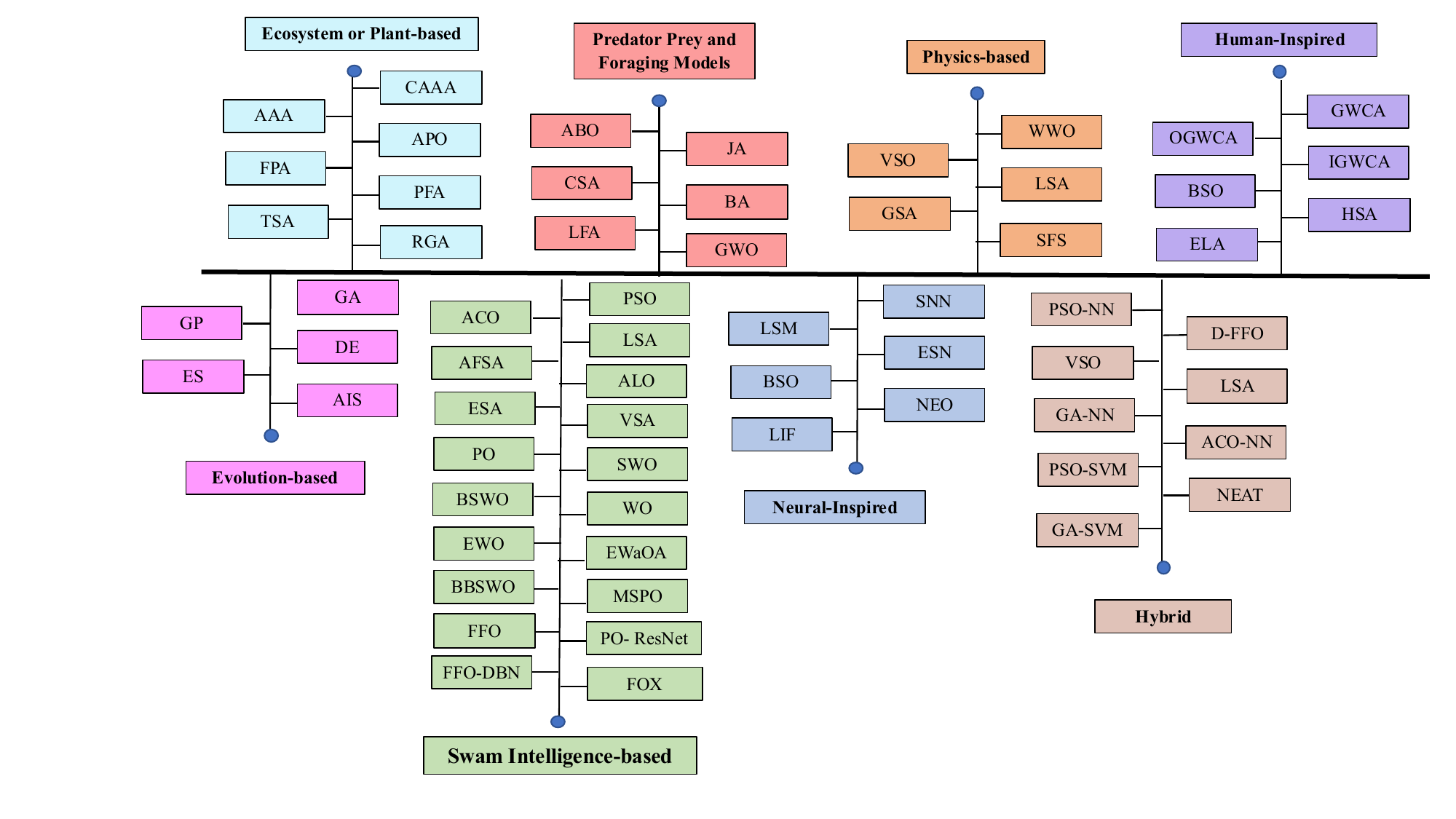}
    \caption{Taxonomy chart of Bio-Inspired Algorithms classified by biological inspiration sources. Foundational algorithms are highlighted alongside contested variants, illustrating both enduring and disputed contributions.}
    \Description{A chart showing the taxonomy of bio-inspired algorithms classified by their biological inspiration source.}
    \label{fig:Taxonomy_chart}
\end{figure}

A visual overview is presented in Figure~\ref{fig:Taxonomy_chart}, where the main categories are mapped alongside representative algorithms. To avoid conflating maturity with novelty, we explicitly note where methods have been validated through rigorous analysis (e.g., GA, ES, DE, PSO, ACO, ABC) and where later proposals (e.g., CS, BA, FA, SSA, GOA) have been critiqued for overlapping with existing paradigms.

\subsection{Evolutionary and Population-Based Algorithms}

The evolutionary family remains the most firmly established category of BIAs. Genetic Algorithms (GA) \cite{Holland1975, goldberg1989genetic}, Genetic Programming (GP) \cite{koza1992genetic, banzhaf1998genetic}, Evolution Strategies (ES) \cite{rechenberg1973evolutionsstrategie, beyer2002evolution}, and Differential Evolution (DE) \cite{storn1997differential} are widely regarded as methodological cornerstones. Their operators—selection, mutation, crossover, self-adaptation, and differencing—introduced genuine innovations that distinguish them from later metaphorical solvers. DE, in particular, remains a critical baseline due to its simple yet powerful differencing operator, while ES is mathematically grounded in self-adaptation theory. These models continue to inform both theoretical work and practical applications, in stark contrast to later “evolution-inspired” solvers that repackage similar mechanics under new analogies \cite{weyland2010rigorous, camacho2023exposing}.

\subsection{Swarm Intelligence Algorithms}

Swarm intelligence introduced a second validated pillar of BIAs. Particle Swarm Optimization (PSO) \cite{kennedy1995particle, clerc2002particle} formalized cognitive–social learning through velocity updates, and Ant Colony Optimization (ACO) \cite{dorigo1996ant, dorigo2004ant} pioneered stigmergic communication for discrete optimization. Both remain heavily benchmarked and conceptually distinct. Artificial Bee Colony (ABC) \cite{karaboga2008performance, karaboga2009survey} also retains recognition as a legitimate swarm model.  

By contrast, later swarm variants such as Firefly Algorithm (FA) \cite{yang2008nature}, Bat Algorithm (BA) \cite{yang2010bat}, Grey Wolf Optimizer (GWO) \cite{mirjalili2014grey}, Salp Swarm Algorithm (SSA) \cite{Mirjalili2017salp}, and Grasshopper Optimization (GOA) \cite{harandi2024grasshopper} exemplify the metaphor-driven wave. Empirical and theoretical analyses demonstrate that FA and BA merely restate attraction–movement rules already modeled in PSO and DE, GWO replicates leader–follower dynamics without new principles, SSA suffers from conceptual flaws including non–shift invariance \cite{castelli2022salp}, and GOA reduces to PSO with parameter re-interpretation \cite{camacho2023exposing, harandi2024grasshopper}. While these methods achieved visibility through metaphorical appeal, their operator-level novelty remains contested.

\subsection{Plant-, Ecosystem-, and Foraging-Inspired Algorithms}

Plant-inspired methods such as Flower Pollination Algorithm (FPA) \cite{yang2012fpa}, Paddy Field Algorithm (PFA) \cite{premaratne2009biologically}, and Algae-based models \cite{turkoglu_chaotic_2025} attempted to translate biological cycles into optimization operators. Ecosystem and foraging-based methods include Bacterial Foraging Optimization (BFO) \cite{passino2002biomimicry}, Cuckoo Search (CS) \cite{yang2009cuckoo}, and Ant Lion Optimization (ALO) \cite{mirjalili2014grey}. While BFO retains niche relevance, most others have been shown to be conceptually redundant: CS is functionally equivalent to DE/ES \cite{camacho2023exposing}, and Harmony Search has been formally proven to reduce to existing evolutionary strategies \cite{weyland2010rigorous}. Reviews consistently highlight that these categories illustrate the overextension of metaphors, with limited theoretical grounding or reproducibility \cite{fister2016new, molina_comprehensive_2020}.

\subsection{Physics- and Human-Inspired Algorithms}

Algorithms based on physical metaphors (e.g., Gravitational Search (GSA) \cite{rashedi2009gsa}, Water Wave Optimization (WWO) \cite{zheng2010wwo}, Vortex Search (VSA) \cite{civicioglu2013vsa}) attempt to model natural forces and dynamics. Similarly, human-inspired methods (e.g., Brain Storm Optimization (BSO) \cite{xue2012brainstorm}, Great Wall Construction Algorithm (GWCA) \cite{guan2023great}) draw from cognitive or cultural processes. While creative, most remain underexplored, insufficiently benchmarked, and often reducible to previously established operators \cite{camacho2023exposing, aranha2022metaphor}. These approaches highlight the risk of conflating metaphorical novelty with algorithmic contribution.

\subsection{Neural-Inspired and Hybrid Models}

A more constructive trajectory is evident in neural-inspired and hybrid methods. NeuroEvolution of Augmenting Topologies (NEAT) \cite{lang2021neat}, PSO–NN hybrids \cite{rauf2018pso_ann}, and GA–PSO or GWO–CSA combinations \cite{ou2023improved, pham2023bio} have demonstrated practical impact in feature selection, classification, and reinforcement learning. Unlike metaphor-driven solvers, these approaches generate value through integration—leveraging complementary mechanisms to improve convergence, adaptability, and scalability. This direction reflects the pragmatic shift observed in recent years, where research emphasizes cross-paradigm hybridization and domain-specific tuning rather than entirely new metaphors \cite{das2024hybrid, nadimi2023systematic}.

\subsection{Critical Synthesis}

Overall, taxonomy reveals two parallel streams in the evolution of BIAs. Foundational categories—evolutionary computation (GA, ES, DE) and swarm intelligence (PSO, ACO, ABC)—introduced operators that remain validated and enduring. In contrast, plant-, foraging-, physics-, and human-inspired algorithms largely illustrate the metaphor-driven proliferation critiqued by recent reviews \cite{sorensen2015metaheuristics, camacho2023exposing}. Hybrid and neural-inspired models represent a more constructive trend, where novelty lies not in analogies but in integration and performance-driven adaptation. Thus, taxonomy serves not only as a descriptive classification but also as a critical map of where the field has matured and where it has stagnated.

\section{Algorithmic Overview and Working Principles}
\label{sec:working}

Bio-Inspired Algorithms (BIAs) employ stochastic, population-based heuristics inspired by natural, physical, or social processes. Unlike deterministic solvers, BIAs operate without gradient information, instead iteratively refining populations using operators designed to balance exploration and exploitation. Their appeal lies in adaptability, modularity, and robustness for black-box optimization. Yet, while a handful of foundational models introduced genuinely novel operators supported by theory, many subsequent variants have been critiqued as metaphor-driven restatements of existing mechanisms \cite{camacho2023exposing, aranha2022metaphor}. This section reviews the main algorithmic families—evolutionary algorithms, swarm intelligence, and other nature-inspired variants—emphasizing operator mechanics, theoretical contributions, and critical commentary.

\subsection{Evolutionary Algorithms}

Evolutionary algorithms (EAs) were among the first population-based BIAs and remain some of the most validated. They model Darwinian principles of inheritance, variation, and survival of the fittest. Their operators—mutation, recombination, and selection—have been rigorously analyzed through schema theory, Markov chains, and stochastic process models \cite{goldberg1989genetic, vose1999simple}.

\textbf{Genetic Algorithm (GA)} \cite{Holland1975}: uses crossover and mutation to explore the search space. Mutation is often expressed as Gaussian perturbation:
\begin{equation}
x'_i = x_i + \delta, \quad \delta \sim \mathcal{N}(0, \sigma^2).
\end{equation}
GA remains foundational, with proven theoretical models and widespread applications in scheduling, feature selection, and engineering optimization.

\textbf{Genetic Programming (GP)} \cite{koza1992genetic, banzhaf1998genetic}: evolves symbolic tree-structured programs using subtree crossover and mutation. GP is valued for producing interpretable models in regression, classification, and control.

\textbf{Evolution Strategies (ES)} \cite{rechenberg1973evolutionsstrategie, beyer2002evolution}: emphasize self-adaptive mutation step sizes. A candidate solution mutates as
\begin{equation}
x' = x + \mathcal{N}(0, \sigma^2),
\end{equation}
with $\sigma$ evolving dynamically. ES are mathematically grounded and reliable in high-dimensional continuous spaces.

\textbf{Differential Evolution (DE)} \cite{storn1997differential}: introduces mutation-by-differencing,
\begin{equation}
v_i = x_{r1} + F \cdot (x_{r2} - x_{r3}),
\end{equation}
with $F$ as a scaling factor. DE remains one of the most efficient solvers for continuous optimization and is widely recognized as a genuine methodological innovation.

In sum, GA, GP, ES, and DE established the operator-level foundations of evolutionary computation. Later “evolution-inspired” solvers such as Cuckoo Search and Bat Algorithm are largely reformulations of these mechanisms under new analogies \cite{camacho2023exposing}.

\subsection{Swarm Intelligence Algorithms}

Swarm Intelligence (SI) algorithms are based on decentralized cooperation among simple agents. The most influential methods remain PSO, ACO, and ABC; later SI variants are widely cited but rarely introduce operator-level novelty.

\textbf{Particle Swarm Optimization (PSO)} \cite{kennedy1995particle}: updates velocity and position as
\begin{equation}
v_i^{t+1} = w v_i^t + c_1 r_1 (p_i - x_i^t) + c_2 r_2 (g - x_i^t),
\end{equation}
\begin{equation}
x_i^{t+1} = x_i^t + v_i^{t+1}.
\end{equation}
Convergence has been studied through eigenvalue and Lyapunov analysis \cite{clerc2002particle}. PSO remains a baseline for continuous optimization and parameter tuning.

\textbf{Ant Colony Optimization (ACO)} \cite{dorigo1996ant, dorigo2004ant}: constructs probabilistic paths using pheromone intensity,
\begin{equation}
P_{ij}(t) = \frac{[\tau_{ij}(t)]^\alpha [\eta_{ij}]^\beta}{\sum_{k \in N_i} [\tau_{ik}(t)]^\alpha [\eta_{ik}]^\beta},
\end{equation}
and has been rigorously validated for routing and scheduling \cite{blum2005ant}.

\textbf{Artificial Bee Colony (ABC)} \cite{karaboga2008performance, karaboga2009survey}: employs roles for employed, onlooker, and scout bees. A common update rule is
\begin{equation}
v_{ij} = x_{ij} + \phi_{ij}(x_{ij} - x_{kj}), \quad \phi_{ij} \in [-1,1].
\end{equation}
ABC has proven effective in clustering and engineering design, though it is less theoretically developed than PSO or ACO.

\textbf{Contested Variants:} Firefly Algorithm (FA) \cite{yang2008nature}, Bat Algorithm (BA) \cite{yang2010bat}, Grey Wolf Optimizer (GWO) \cite{mirjalili2014grey}, Whale Optimization Algorithm (WOA) \cite{mirjalili2016whale}, Dragonfly Algorithm (DA) \cite{Mirjalili2016}, Salp Swarm Algorithm (SSA) \cite{Mirjalili2017salp}, and Grasshopper Optimization (GOA) \cite{harandi2024grasshopper} illustrate the metaphor-driven expansion. Subsequent analyses show that their mechanics—attraction, spiral motion, or leader–follower dynamics—are restatements of PSO or DE principles, often with weaker performance or even conceptual flaws \cite{castelli2022salp, camacho2023exposing}.

\subsection{Other Nature-Inspired Variants}

Beyond evolutionary and swarm models, other metaphors include foraging, plant biology, and physics. While diverse in framing, most overlap heavily with established operators.

\textbf{Foraging Algorithms:} Bacterial Foraging Optimization (BFO) \cite{passino2002biomimicry} uses chemotaxis:
\begin{equation}
\theta_i(t+1) = \theta_i(t) + C(i) \cdot \frac{\Delta(i)}{\|\Delta(i)\|}.
\end{equation}
It is noise-robust but computationally expensive. Cuckoo Search (CS) \cite{yang2009cuckoo}, despite its popularity, has been shown equivalent to DE/ES \cite{camacho2023exposing}.

\textbf{Plant-Inspired Algorithms:} Flower Pollination Algorithm (FPA) \cite{yang2012fpa} and Paddy Field Algorithm (PFA) \cite{premaratne2009biologically} employ pollination and agricultural analogies but often reduce to Lévy flights or recombination \cite{fister2016new}.

\textbf{Physics-Inspired Algorithms:} Gravitational Search (GSA) \cite{rashedi2009gsa}, Water Wave Optimization (WWO) \cite{zheng2010wwo}, and Vortex Search (VSA) \cite{civicioglu2013vsa} adopt attraction–oscillation dynamics already embedded in PSO/DE, and are generally regarded as contested variants \cite{sorensen2015metaheuristics}.

\begingroup
\fontsize{8pt}{9pt}\selectfont
\begin{longtable}{m{0.22\linewidth} m{0.28\linewidth} m{0.14\linewidth} m{0.28\linewidth}}
\caption{Comparison of representative Bio-Inspired Algorithms (BIAs), summarizing core operators, validation status, and critical commentary.}
\label{tab:algorithmic_comparison} \\
\toprule
\textbf{Algorithm (with source)} & \textbf{Core Operator / Equation} & \textbf{Validation Status} & \textbf{Critical Notes} \\
\midrule
\endfirsthead
\toprule
\textbf{Algorithm (with source)} & \textbf{Core Operator / Equation} & \textbf{Validation Status} & \textbf{Critical Notes} \\
\midrule
\endhead
\midrule
\endfoot
\bottomrule
\endlastfoot

\textbf{Genetic Algorithm (GA)} \cite{Holland1975} & Mutation: $x'_i = x_i + \delta,\; \delta \sim \mathcal{N}(0,\sigma^2)$ & Foundational & Supported by schema theory and Markov chain models; enduring benchmark across domains \cite{goldberg1989genetic, vose1999simple}. \\
\midrule
\textbf{Genetic Programming (GP)} \cite{koza1992genetic} & Subtree crossover and mutation on symbolic trees & Foundational & Legitimate extension of GA; robust applications in regression and control \cite{banzhaf1998genetic}. \\
\midrule
\textbf{Evolution Strategies (ES)} \cite{rechenberg1973evolutionsstrategie} & Self-adaptive Gaussian mutation: $x' = x + \mathcal{N}(0,\sigma^2)$ & Foundational & Strong theoretical grounding; convergence proofs in continuous optimization \cite{beyer2002evolution}. \\
\midrule
\textbf{Differential Evolution (DE)} \cite{storn1997differential} & Mutation by differencing: $v_i = x_{r1} + F(x_{r2}-x_{r3})$ & Foundational & Introduced genuine novelty; many later solvers (e.g., CS) reduce to DE variants \cite{camacho2023exposing}. \\
\midrule
\textbf{PSO, ACO, ABC} \cite{kennedy1995particle, dorigo1996ant, karaboga2008performance} & Social learning (PSO), pheromone trails (ACO), neighbor exploitation (ABC) & Foundational & Validated swarm models; theoretical analyses via Lyapunov (PSO) and Markov chains (ACO). \\
\midrule
\textbf{Contested Swarm/Ecological Variants:} FA \cite{yang2008nature}, BA \cite{yang2010bat}, GWO \cite{mirjalili2014grey}, WOA \cite{mirjalili2016whale}, DA \cite{Mirjalili2016}, SSA \cite{Mirjalili2017salp}, GOA \cite{harandi2024grasshopper} & Attraction decay, frequency/loudness updates, leader–follower encircling, spiral motion, alignment/cohesion, chain following, attraction–repulsion & Contested & Operators overlap with PSO/DE; novelty disputed; SSA shown non–shift-invariant and underperforms random search \cite{camacho2023exposing, castelli2022salp}. \\
\midrule
\textbf{Bacterial Foraging Optimization (BFO)} \cite{passino2002biomimicry} & Chemotaxis: $\theta_{i}(t+1)=\theta_{i}(t)+C(i)\frac{\Delta(i)}{\|\Delta(i)\|}$ & Contested but niche & Robust under noise; limited scalability and runtime inefficiency \cite{das2009bacterial}. \\
\midrule
\textbf{Plant-Inspired Algorithms:} FPA \cite{yang2012fpa}, PFA \cite{premaratne2009biologically}, APOA \cite{zhao2011apoa} & Pollination/growth analogies, often Lévy flights & Contested & Rephrase random walk/recombination; rarely benchmarked \cite{fister2016new}. \\
\midrule
\textbf{Physics-Inspired Algorithms:} GSA \cite{rashedi2009gsa}, WWO \cite{zheng2010wwo}, VSA \cite{civicioglu2013vsa} & Attraction/oscillation analogies & Contested & Dynamics mimic PSO/DE; weak theoretical justification \cite{sorensen2015metaheuristics}. \\
\midrule
\textbf{BSO, NEAT, Hybrid Models} \cite{xue2012brainstorm, lang2021neat, pham2023bio, das2024hybrid} & Cluster recombination; topology/weight evolution; cross-paradigm integration & Emerging / Constructive & Constructive novelty lies in hybridization with ML; promising results in feature selection and reinforcement learning. \\
\end{longtable}
\endgroup

\subsection{Human-Inspired, Neural-Inspired, and Hybrid Models}

Human-inspired solvers (e.g., Brain Storm Optimization (BSO) \cite{xue2012brainstorm}) replicate social or cognitive processes but remain insufficiently benchmarked. By contrast, neural-inspired and hybrid models have demonstrated constructive novelty. NeuroEvolution of Augmenting Topologies (NEAT) \cite{lang2021neat} evolves neural architectures; hybrid methods (e.g., PSO–NN \cite{rauf2018pso_ann}, GA–SVM \cite{lee2021ga_dl}, ACO–NN \cite{mavrovouniotis2014aco}) integrate BIAs with machine learning, yielding effective tools for feature selection, reinforcement learning, and hyperparameter tuning \cite{pham2023bio, das2024hybrid}. These contributions highlight that practical advances often arise from integration rather than metaphors.

\subsection{Meta-Analysis: Rise and Fall of BIAs}

The trajectory of BIAs reflects both genuine innovation and metaphor-driven oversaturation. Foundational models (GA, ES, DE, PSO, ACO, ABC) introduced enduring operators with theoretical justification, while later metaphor-driven solvers (e.g., FA, BA, SSA) largely recycled these ideas with limited novelty \cite{sorensen2015metaheuristics, camacho2023exposing}. As summarized in Table~\ref{tab:algorithmic_comparison} and Table~\ref{tab:bias_capabilities}, a small subset of algorithms continues to provide durable value, while most contested variants have failed to demonstrate general-purpose utility. The field’s long-term lesson is that creative metaphors can stimulate innovation but, without methodological rigor, lead to proliferation without progress.

\begingroup
\fontsize{8pt}{9pt}\selectfont
\begin{longtable}{m{0.23\linewidth} m{0.20\linewidth} m{0.22\linewidth} m{0.25\linewidth}}
\caption{Classification of BIAs by core capabilities, domains, and critical remarks.}
\label{tab:bias_capabilities} \\
\toprule
\textbf{Algorithm (with source)} & \textbf{Core Capability} & \textbf{Problem Domains} & \textbf{Remarks} \\
\midrule
\endfirsthead
\toprule
\textbf{Algorithm (with source)} & \textbf{Core Capability} & \textbf{Problem Domains} & \textbf{Remarks} \\
\midrule
\endhead
\midrule
\endfoot
\bottomrule
\endlastfoot

GA, ES \cite{Holland1975, rechenberg1973evolutionsstrategie} & Global exploration, recombination, self-adaptation & Combinatorial optimization, scheduling, parameter tuning & Foundational methods; schema theory and convergence analysis established \cite{goldberg1989genetic, beyer2002evolution}. \\
\midrule
DE \cite{storn1997differential} & Self-scaling mutation for continuous search & Engineering design, control, continuous optimization & Highly efficient; many later BIAs reduce to DE variants \cite{camacho2023exposing}. \\
\midrule
PSO, ACO, ABC \cite{kennedy1995particle, dorigo1996ant, karaboga2008performance} & Social/pheromone-based exploration–exploitation & Robotics, ML hyperparameters, routing, clustering, scheduling & Strong theoretical support; widely benchmarked in practice. \\
\midrule
Contested swarm/ecological group: FA, BA, GWO, WOA, DA, SSA, GOA \cite{yang2008nature, yang2010bat, mirjalili2014grey, mirjalili2016whale, Mirjalili2016, Mirjalili2017salp, harandi2024grasshopper} & Attraction/perturbation, leader–follower dynamics, spiral/chain motions & Claimed across engineering, ML, and bioinformatics & Operators equivalent to PSO/DE; novelty disputed; SSA underperforms random search \cite{camacho2023exposing, castelli2022salp}. \\
\midrule
BFO \cite{passino2002biomimicry} & Noise-robust chemotaxis & Control, signal processing & Effective in noisy domains but poor scalability \cite{das2009bacterial}. \\
\midrule
Plant-inspired: FPA, PFA, APOA \cite{yang2012fpa, premaratne2009biologically, zhao2011apoa} & Pollination/growth analogies & Scheduling, multi-objective optimization & Limited novelty; rarely benchmarked \cite{fister2016new}. \\
\midrule
Physics-inspired: GSA, WWO, VSA \cite{rashedi2009gsa, zheng2010wwo, civicioglu2013vsa} & Attraction/oscillation analogies & Structural optimization, engineering & Overlaps with PSO/DE; novelty questioned \cite{sorensen2015metaheuristics}. \\
\midrule
BSO, NEAT, Hybrid Models \cite{xue2012brainstorm, lang2021neat, pham2023bio, das2024hybrid} & Cognitive clustering; neural evolution; ML integration & Optimization, feature selection, RL, classification & Constructive direction; novelty lies in integration rather than metaphor. \\
\end{longtable}
\endgroup


\section{Applications Across Domains} 
\label{sec:applications}

Bio-Inspired Algorithms (BIAs) have been widely applied across multiple disciplines due to their ability to handle nonlinear, multimodal, and dynamic optimization problems. However, it is important to distinguish between well-validated applications, primarily relying on foundational algorithms such as GA, DE, PSO, and ACO, and more speculative claims based on metaphor-driven variants such as GWO, WOA, FA, and SSA. This section provides a critical review of applications in five broad domains-engineering, computer science, robotics, bioinformatics, and networking-while highlighting where evidence is strong and where further validation is required.

\subsection{Engineering}
Engineering design and control problems often involve large-scale, nonlinear, and multi-objective formulations. Foundational BIAs such as GA, PSO, and DE remain the dominant methods for structural optimization, PID control, and energy dispatch \citep{zakariyya2023optimal, faris2018grey, ajlouni2025enhancing}. For instance, GA and DE have been extensively validated for topology optimization of trusses and frames, and hybrid GA–PSO variants (GAPSO) provide robust solutions for power system dispatch under uncertainty \citep{maqbool2025hybrid}. By contrast, newer metaphoric algorithms like GWO or WOA are often reported in structural or energy engineering contexts \citep{faris2018grey}, but critical evaluations show their operators overlap with PSO/DE, raising questions about true novelty and reproducibility \citep{camacho2023exposing, castelli2022salp}. 

\subsection{Computer Science}
In computer science, BIAs are used in feature selection, scheduling, and image processing. PSO and GA dominate feature selection due to their balance of exploration and exploitation, while ABC and ACO are applied to clustering and task scheduling \citep{murphy2022particle, tariqcomparative}. Hybrid models (e.g., GAPSO, WOA–PSO) show promise in cloud and edge scheduling \citep{nadimi2023systematic}, yet again the novelty lies more in hybridization than in the base metaphoric algorithm. Algorithms such as FA and GWO are sometimes applied in image processing \citep{zakariyya2023optimal}, but evidence for their consistent superiority over PSO/GA remains limited, with many applications relying on problem-specific parameter tuning.

\subsection{Robotics}
Robotics applications emphasize path planning, navigation, and distributed control. Classical swarm methods such as ACO and PSO have been robustly validated for navigation and obstacle avoidance \citep{li2019survey, poy2024enhanced}. GWO and ABC have also been explored for swarm robotics tasks such as formation control \citep{li2021bio}, but again these often replicate PSO-like dynamics with additional metaphorical framing. More credible progress in robotics arises from hybrid approaches that integrate BIAs with controllers such as neural networks or backstepping methods, where the optimization component is grounded in well-understood algorithms like PSO or GA.

\subsection{Bioinformatics}
In bioinformatics, BIA applications are more recent but particularly high-impact due to the scale of genomic data. Binary PSO variants (BPSO, EBPSO) are widely used for gene selection, providing compact feature subsets with strong predictive performance \citep{murphy2022particle}. GA and ACO have been applied to sequence alignment and protein folding \citep{zakariyya2023optimal}, while Firefly and hybrid algorithms have been reported in biomarker discovery. Here again, EBPSO and GA-based hybrids appear reproducible, but applications of FA or WOA remain under-validated and often lack cross-dataset benchmarking.

\subsection{Networking}
Networking applications often exploit graph-based optimization. ACO is the most consistently validated method for routing in MANETs and WSNs, including extensions like quantum-inspired ACO (QACO) \citep{khudair2023quantum}. PSO and GA are applied to resource allocation and load balancing in IoT and cloud networks \citep{nadimi2023systematic}. WOA and GWO appear in networking literature but without rigorous benchmarking against established baselines, often serving more as illustrative case studies than as mainstream solutions.

\begin{landscape}
\thispagestyle{empty}
\begingroup
\fontsize{8pt}{9pt}\selectfont
\begin{table}[htbp]
  \centering
  \caption{Critical category-based summary of Bio-Inspired Algorithms (BIAs), highlighting inspirations, mechanisms, strengths, and limitations.}
  \label{tab:comprehensive_category}
  \begin{tabular}{p{1.5cm} p{3.3cm} p{3.3cm} p{4.3cm} p{4.2cm} p{2.2cm}}
    \toprule
    \textbf{Category} & \textbf{Inspiration} & \textbf{Core Mechanism} & \textbf{Strengths (Validated)} & \textbf{Limitations (Critiques)} & \textbf{Representative Variants} \\
    \midrule

    Swarm Intelligence-Based 
    \citep{dorigo1996ant, kennedy1995particle, karaboga2008performance, nadimi-shahraki_mfo-sfr_2023} &
    Social foraging, pheromone trails, collective movement (ants, birds, bees, moths). &
    Leader–follower roles, adaptive foraging, and probabilistic search strategies. &
    Well-validated: ACO (routing/scheduling), PSO (continuous optimization). Extensively benchmarked and scalable. &
    Later metaphors (MFO, GWO, FA) shown to overlap with PSO/DE; novelty disputed \citep{camacho2023exposing, aranha2022metaphor}. &
    ACO, PSO, ABC, MFO, GWO, FA, SSA \\

    \midrule

    Physics-Based Algorithms 
    \citep{rashedi2009gsa, zheng2010wwo, civicioglu2013vsa, abualigah_lightning_2021} &
    Gravitational attraction, wave propagation, vortex dynamics. &
    Energy propagation, oscillatory updates, attraction–repulsion models. &
    Found some use in structural optimization and energy systems. &
    Often reformulations of PSO/DE; limited cross-domain validation; criticized as metaphor-driven \citep{sorensen2015metaheuristics}. &
    GSA, WWO, VSA, Lightning Search \\

    \midrule

    Ecosystem / Plant-Based 
    \citep{yang2012fpa, premaratne2009biologically, zhao2011apoa, turkoglu_chaotic_2025} &
    Pollination, algae growth, seasonal farming cycles. &
    Lévy flights, branching/growth heuristics, chaotic maps for diversity. &
    Some niche use in scheduling and clustering; chaos-enhanced versions improve exploration. &
    Weak empirical validation; performance often indistinguishable from random walk or recombination \citep{fister2016new}. &
    FPA, APOA, AAA, CAAA, PFA \\

    \midrule

    Predator–Prey / Foraging 
    \citep{passino2002biomimicry, odili_stochastic_2022, yang_dynamic_2018} &
    Hunting and survival dynamics (wolves, lions, bacteria, fish). &
    Role-based leader–follower models, chemotaxis, adaptive hunting trajectories. &
    BFO validated in noisy control problems; ACO remains strong in routing. &
    Most new metaphors (GWO, WOA, SSA) criticized as PSO/DE variants with weak novelty \citep{castelli2022salp, camacho2023exposing}. &
    BFO, GWO, WOA, SSA, ALO, AFSA \\

    \midrule

    Human/Culture-Inspired 
    \citep{xue2012brainstorm, guan2023great, zeng2025assignment} &
    Cognitive brainstorming, social hierarchies, human collaboration. &
    Cluster-based recombination, role switching, opposition-based learning. &
    BSO shows interesting cognitive metaphor for feature selection. &
    Many others remain problem-specific with limited reproducibility; generalization lacking. &
    BSO, GWCA, IGWCA, OGWCA \\

    \midrule

    Hybrid Algorithms 
    \citep{pham2023bio, osaba2021tutorial, das2024hybrid} &
    Integration of GA/PSO/ACO with ML models (ANN, SVM, DL). &
    Uses evolutionary/swarm search to optimize ML hyperparameters or architectures. &
    Demonstrated effectiveness in healthcare, cybersecurity, feature selection. &
    High computational cost; novelty lies in integration, not in metaphor. &
    PSO–NN, GA–SVM, ACO–NN, NEAT \\
    
    \bottomrule
  \end{tabular}
\end{table}
\endgroup
\end{landscape}

\subsection*{Discussion on Applications}
Table~\ref{tab:comprehensive_category} critically summarizes BIA categories by strengths and limitations, illustrating the frequent gap between claimed benefits and empirical validation. Table~\ref{tab:application_domains} complements this by mapping specific algorithms to domains, making clear that while PSO, GA, DE, and ACO underpin most validated applications, metaphor-driven variants rarely demonstrate domain-specific superiority. Taken together, the evidence indicates that the long-term impact of BIAs lies in a handful of foundational methods and their hybrid extensions, while the majority of metaphor-inspired variants remain peripheral.

\thispagestyle{empty}
\begingroup
\fontsize{8pt}{9pt}\selectfont
\begin{table}[h!]
  \centering
  \caption{Validated applications of Bio-Inspired Algorithms (BIAs) across domains. Only well-documented cases are shown; metaphor-driven variants with limited validation are excluded or noted with caution.}
  \label{tab:application_domains}
  \begin{tabular}{p{1.1cm} p{2.2cm} p{1.8cm} p{2.2cm} p{2cm} p{2.2cm}}
    \toprule
    \textbf{Algorithm} & \textbf{Engineering} & \textbf{Computer Science} & \textbf{Robotics} & \textbf{Bioinformatics} & \textbf{Networking} \\
    \midrule
    PSO & Structural/topology optimization; PID tuning \citep{zakariyya2023optimal, faris2018grey} & Feature selection, scheduling \citep{murphy2022particle, tariqcomparative} & Path planning, obstacle avoidance \citep{poy2024enhanced, ou2023improved} & Gene signature discovery \citep{murphy2022particle} & WSN clustering, load balancing \citep{zakariyya2023optimal} \\
    \midrule
    GA & Power flow, PID control \citep{ajlouni2025enhancing} & Data mining, optimization \citep{tariqcomparative} & Adaptive controllers \citep{ajlouni2025enhancing} & Biomarker discovery \citep{murphy2022particle} & VM scheduling in cloud \citep{tariqcomparative} \\
    \midrule
    DE & Structural design, energy dispatch \citep{maqbool2025hybrid} & Continuous optimization \citep{osaba2021tutorial} & – & Gene/protein modeling (limited) & Resource allocation in IoT/cloud \citep{nadimi2023systematic} \\
    \midrule
    ACO & Load optimization \citep{zakariyya2023optimal} & Job scheduling \citep{zakariyya2023optimal} & Navigation and path planning \citep{li2019survey} & Sequence alignment, protein folding \citep{zakariyya2023optimal} & Routing in MANETs, QACO for gateway discovery \citep{khudair2023quantum} \\
    \midrule
    ABC & Signal design \citep{zakariyya2023optimal} & Image processing, clustering \citep{zakariyya2023optimal} & Cooperative multi-robot control \citep{li2021bio} & Cancer biomarker detection \citep{zakariyya2023optimal} & Bandwidth allocation in virtual networks \citep{zakariyya2023optimal} \\
    \midrule
    GWO (contested) & Applied in energy dispatch \citep{faris2018grey} & Image segmentation \citep{faris2018grey} & Formation control \citep{li2021bio} & – & Reported in IoT/cloud \citep{nadimi2023systematic}, though novelty is disputed \citep{camacho2023exposing} \\
    \midrule
    Hybrid (e.g., GAPSO, PSO–NN) & Renewable energy dispatch \citep{maqbool2025hybrid} & Cloud/edge optimization \citep{tariqcomparative} & PID tuning \citep{ajlouni2025enhancing} & Feature selection in genomics \citep{pham2023bio} & VM mapping, IoT scheduling \citep{nadimi2023systematic} \\
    \bottomrule
  \end{tabular}
\end{table}
\endgroup


\section{Benchmarking and Comparative Analysis} 
\label{sec:benchmarking}

Benchmarking plays a central role in separating well-validated algorithms from contested or weakly justified ones. It exposes which Bio-Inspired Algorithms (BIAs) remain credible in mainstream optimization and which fail to demonstrate consistent utility. While early algorithms such as GA, DE, PSO, and ACO are repeatedly validated across diverse benchmarks, many metaphor-driven variants (e.g., Firefly, Bat, Grey Wolf, Salp Swarm) show limited novelty and underperform when compared under rigorous test conditions \citep{sorensen2015metaheuristics, camacho2023exposing, castelli2022salp, aranha2022metaphor}. This section critically reviews performance metrics, comparative results, benchmark functions, and emerging hybridization trends to highlight the rise, saturation, and decline of BIAs in optimization research.

\subsection{Performance Metrics}

To enable fair comparison, several standardized performance metrics have been established. These include convergence speed, robustness, memory use, computational complexity, and scalability. Convergence speed evaluates how rapidly an algorithm approaches optimality, which is crucial for real-time systems. Robustness captures consistency across multiple runs under noise or dynamic conditions. Memory use and computational complexity reflect efficiency in resource-constrained environments. Scalability measures how algorithms sustain performance with increased dimensionality or data size. 

Importantly, these metrics have revealed a key pattern: foundational BIAs (GA, DE, PSO, ACO) maintain strong performance across dimensions and domains, while newer metaphor-driven BIAs often collapse under rigorous benchmarking, showing no improvement beyond random search or simple baselines \citep{camacho2023exposing, castelli2022salp}. Thus, benchmarking has become not just a tool for selection, but a filter distinguishing genuine methodological contributions from superficial metaphors.

\subsection{Comparative Evaluations}

Comparative studies consistently demonstrate the durability of early, theoretically grounded algorithms and the fragility of later metaphor-driven ones. For example, PSO remains widely adopted due to its simple update rules and proven scalability, while DE has become a gold standard in continuous optimization because of its efficient mutation and recombination mechanisms. GA and ACO, though older, continue to be robust performers in discrete, combinatorial, and multi-objective settings.

By contrast, algorithms such as FA, BA, SSA, and GWO rarely outperform DE or PSO under identical conditions and often reduce to minor variations of existing operators \citep{camacho2023exposing, aranha2022metaphor}. As highlighted in Table~\ref{tab:bia_comparison}, the distinction between foundational and contested BIAs is stark: while the former remain benchmarks in academia and industry, the latter persist mainly in literature with limited validation or adoption. Hybrid approaches appear as a middle ground, leveraging operator reuse from validated methods while discarding weak metaphorical elements.

\subsection{Benchmark Functions and Datasets}

Standardized benchmarks have played a pivotal role in these evaluations. Classical continuous functions such as Rastrigin, Rosenbrock, Ackley, Sphere, and Griewank allow reproducible comparisons of exploration–exploitation trade-offs and convergence behavior. In combinatorial optimization, problems such as the Travelling Salesman Problem (TSP) remain canonical for testing search efficiency and solution quality. 

Datasets such as those from the UCI Machine Learning Repository \citep{Dua:2019}, The Cancer Genome Atlas (TCGA) \citep{NCI_TCGA}, and cloud simulation frameworks like CloudSim \citep{calheiros2011cloudsim} provide more realistic testbeds, exposing how algorithms scale in high-dimensional, noisy, or domain-specific contexts. These benchmarks consistently reaffirm the strength of GA, DE, PSO, and ACO, while newer metaphors often fail to transfer beyond toy problems. As noted by Osaba et al. \citep{osaba2021tutorial}, the credibility of BIAs depends not only on novel inspiration but on consistent reproducibility across such standardized settings.

\thispagestyle{empty}
\begingroup
\fontsize{8pt}{9pt}\selectfont
\begin{longtable}{m{0.14\linewidth} m{0.26\linewidth} m{0.15\linewidth} m{0.35\linewidth}}
\caption{Comparison of representative Bio-Inspired Algorithms (BIAs), summarizing their core operators, validation status, and critical commentary.}
\label{tab:bia_comparison} \\
\toprule
\textbf{Algorithm} & \textbf{Core Operator / Equation} & \textbf{Validation Status} & \textbf{Critical Notes} \\
\midrule
\endfirsthead
\toprule
\textbf{Algorithm} & \textbf{Core Operator / Equation} & \textbf{Validation Status} & \textbf{Critical Notes} \\
\midrule
\endhead
\midrule
\endfoot
\bottomrule
\endlastfoot

\textbf{Genetic Algorithm (GA)} \cite{Holland1975, goldberg1989genetic} & Mutation: $x'_i = x_i + \delta, \; \delta \sim \mathcal{N}(0,\sigma^2)$ & Foundational & Rigorously analyzed via schema theory and Markov models; validated across domains. \\

\textbf{Genetic Programming (GP)} \cite{koza1992genetic, banzhaf1998genetic} & Subtree crossover, mutation on symbolic trees & Foundational & Legitimate extension of GA for symbolic regression; strong applications in interpretable ML. \\

\textbf{Evolution Strategies (ES)} \cite{rechenberg1973evolutionsstrategie, beyer2002evolution} & Self-adaptive mutation: $x' = x + \mathcal{N}(0,\sigma^2)$ & Foundational & Mathematically grounded; effective in high-dimensional real-valued optimization. \\

\textbf{Differential Evolution (DE)} \cite{storn1997differential} & Mutation by differencing: $v_i = x_{r1} + F(x_{r2}-x_{r3})$ & Foundational & Highly efficient; considered a baseline in continuous optimization. \\

\textbf{Particle Swarm Optimization (PSO)} \cite{kennedy1995particle, clerc2002particle} & Velocity update with inertia + cognitive/social terms & Foundational & Widely validated; convergence studied via Lyapunov stability. \\

\textbf{Ant Colony Optimization (ACO)} \cite{dorigo1996ant, blum2005ant} & Probabilistic path construction: $P_{ij}(t)$ via pheromone intensity & Foundational & Strong theoretical basis; effective in routing, scheduling, and combinatorial tasks. \\

\textbf{Artificial Bee Colony (ABC)} \cite{karaboga2008performance, karaboga2009survey} & Neighbor exploitation: $v_{ij} = x_{ij} + \phi_{ij}(x_{ij}-x_{kj})$ & Foundational / Applied & Well-studied swarm model; less theoretical maturity than PSO/ACO. \\

\textbf{Firefly Algorithm (FA)} \cite{yang2008nature} & Attraction decay: $\beta(r)=\beta_0 e^{-\gamma r^2}$ & Contested \cite{camacho2023exposing} & Shown equivalent to PSO/DE attraction rules; novelty weak and largely metaphorical. \\

\textbf{Bat Algorithm (BA)} \cite{yang2010bat} & Echolocation updates using loudness and frequency & Contested \cite{camacho2023exposing} & Critiqued as reformulation of GA/PSO; lacks distinct operators. \\

\textbf{Cuckoo Search (CS)} \cite{yang2009cuckoo, yang2010engineering} & Lévy flight exploration: $x_i^{t+1}=x_i^t+\alpha L(\lambda)$ & Contested \cite{camacho2022analysis} & Shown reducible to DE/ES; novelty claims disputed. \\

\textbf{Grey Wolf Optimizer (GWO)} \cite{mirjalili2014grey} & Leader–follower encircling dynamics & Contested \cite{camacho2023exposing} & Overlaps with PSO hierarchy; lacks theoretical grounding. \\

\textbf{Whale Optimization Algorithm (WOA)} \cite{mirjalili2016whale} & Spiral encircling updates & Contested \cite{camacho2023exposing} & Adds metaphorical framing; operator structure similar to existing swarm models. \\

\textbf{Dragonfly Algorithm (DA)} \cite{Mirjalili2016} & Alignment + cohesion + separation dynamics & Contested \cite{camacho2023exposing} & Replicates PSO-like motion rules; weak empirical support. \\

\textbf{Salp Swarm Algorithm (SSA)} \cite{Mirjalili2017salp} & Leader–follower chain propagation & Contested \cite{castelli2022salp} & Demonstrated underperformance vs. random search; lacks operator novelty. \\

\textbf{Grasshopper Optimization (GOA)} \cite{harandi2024grasshopper} & Attraction–repulsion swarming & Contested \cite{camacho2023exposing} & Functionally reducible to PSO; novelty questioned. \\

\textbf{Bacterial Foraging Optimization (BFO)} \cite{passino2002biomimicry, das2009bacterial} & Chemotaxis: $\theta_i(t+1)=\theta_i(t)+C(i)\frac{\Delta(i)}{\|\Delta(i)\|}$ & Contested but niche & Robust in noisy domains; scalability remains limited. \\

\textbf{Plant-Inspired (FPA, PFA, APOA)} \cite{yang2012fpa, premaratne2009biologically, cui2013apoa} & Pollination or growth analogies (often Lévy flights) & Contested & Overlaps with random walk/recombination; weak empirical validation. \\

\textbf{Physics-Inspired (GSA, WWO, VSA)} \cite{rashedi2009gsa, zheng2010wwo, civicioglu2013vsa} & Attraction/oscillation analogies & Contested & Largely metaphorical; operators similar to PSO/DE. \\

\textbf{Brain Storm Optimization (BSO)} \cite{xue2012brainstorm} & Cluster-based recombination and mutation & Emerging & Interesting cognitive metaphor; still limited in benchmarking. \\

\textbf{NeuroEvolution of Augmenting Topologies (NEAT)} \cite{lang2021neat} & Topology + weight evolution & Emerging / Constructive & Genuine novelty; integrates EAs with neural networks. \\

\textbf{Hybrid Models (PSO–NN, GA–SVM, ACO–NN)} \cite{pham2023bio, das2024hybrid} & Cross-paradigm integration & Emerging / Constructive & Strong empirical performance; novelty comes from integration, not metaphor. \\

\end{longtable}
\endgroup

\subsection{Hybridization Trends}

The decline of stand-alone metaphor-driven BIAs has coincided with the rise of hybridization. Hybrid algorithms integrate the strengths of established paradigms (e.g., GA’s genetic operators with PSO’s fast convergence, or DE’s recombination with WOA’s encircling mechanism). These hybrids often outperform their parent algorithms, especially in high-dimensional and dynamic problems, suggesting that the metaphors alone contributed little novelty, while the recombination of operators drives genuine improvement.

Examples include GAPSO, which outperforms GA and PSO individually in load dispatch, PID tuning, and cloud scheduling \citep{maqbool2025hybrid, ajlouni2025enhancing, tariqcomparative}, and pGWO-CSA, which stabilizes Grey Wolf’s weak operator design with CSA’s hypermutation to improve path planning \citep{ou2023improved}. Similarly, hybrid WOA-DE has shown improved convergence in power grid optimization and biomedical signal processing \citep{nadimi2023systematic}. These cases illustrate that hybridization has become the dominant strategy for sustaining relevance in BIA research, effectively marking the end of stand-alone metaphor-driven designs.

Benchmarking thus reveals the broader trajectory of BIAs: from the rise of robust, general-purpose algorithms in the 1970s–1990s, to the oversaturation of metaphor-driven variants in the 2000s–2010s, and finally to the present era of hybridization and integration with machine learning. As Tables~\ref{tab:algorithmic_comparison} and \ref{tab:bia_comparison} illustrate, only a handful of foundational BIAs continue to justify their place as benchmarks, while many others survive only as case studies in the sociology of algorithms. This critical shift underscores that methodological rigor and reproducibility, not metaphorical novelty, determine the lasting value of an optimization algorithm.

\section{Challenges and Research Gaps}
\label{sec:challenges}

Although Bio-Inspired Algorithms (BIAs) achieved widespread popularity and demonstrated versatility across domains, their long-term trajectory reveals a number of critical shortcomings. These unresolved issues not only hinder their broader adoption but also explain why BIAs have declined in mainstream optimization research, being increasingly replaced by better-theorized approaches such as CMA-ES, Bayesian optimization, and hybrid ML-based solvers. This section synthesizes the main challenges that remain open for further investigation and highlights the directions where the field must evolve.

\subsection{Scalability Limitations in High-Dimensional Spaces}

A recurring challenge for BIAs is scalability. While methods such as PSO and ACO perform well on low- to medium-dimensional problems, their performance deteriorates in high-dimensional spaces where the search space grows exponentially. This often results in premature convergence, excessive computational cost, and degraded solution quality \cite{yang_nature-inspired_2020, pham_bio-inspired_2023}. For instance, feature selection problems involving thousands of attributes expose BIAs’ lack of dimensionality awareness, as they typically treat all variables equally without leveraging sparsity or feature relevance \cite{martinez_lights_2021}.  

Attempts at parallelization and hardware acceleration (e.g., FPGA or embedded implementations) have also faced bottlenecks due to memory demands and synchronization overheads \cite{gill_bio-inspired_2019, fu_software_2022}. These limitations explain why BIAs rarely appear in modern large-scale data analytics or real-time systems, where gradient-based deep learning and evolutionary strategies (ES, CMA-ES) dominate. This scalability gap represents both a fundamental weakness and a research opportunity.

\subsection{Weak Theoretical Foundations}

Perhaps the most significant critique of BIAs is their lack of rigorous theoretical grounding. Many algorithms remain heuristic “black boxes,” with convergence proofs or performance guarantees available only for highly idealized settings (e.g., infinite populations, noiseless fitness functions) \cite{poli_particle_2007, sudholt_new_2011}. Although early methods such as GA and PSO were analyzed using schema theory and dynamical systems \cite{goldberg1989genetic, clerc_particle_2002}, most newer metaphor-driven variants lack any formal justification.  

This weakness has two consequences. First, BIAs are often unpredictable and difficult to trust in safety-critical applications such as healthcare or autonomous systems \cite{karaboga_comprehensive_2014}. Second, the unchecked proliferation of metaphor-inspired algorithms-Firefly, Bat, Grey Wolf, Salp Swarm, etc.-has been criticized as methodologically shallow, since many of these variants reduce to minor modifications of GA or PSO under reformulation \cite{sorensen_metaheuristicsmetaphor_2015, camacho2023exposing}. The absence of rigorous theory partly explains the decline of BIAs in mainstream optimization.

\subsection{Parameter Sensitivity and Meta-Optimization}

BIAs are highly sensitive to parameter choices such as mutation rate, inertia weight, or colony size. Optimal configurations are strongly problem-dependent and typically discovered by trial and error, which undermines reproducibility and practical deployment \cite{eiben_parameter_2011}. Poorly tuned parameters can cause premature convergence or instability (e.g., swarm collapse in PSO).  

Recent research on adaptive and self-adaptive parameter control, meta-optimization using Bayesian search or reinforcement learning, and success-history adaptation in DE has improved robustness \cite{tanabe_success-history_2013, hutter_automated_2019, yin_reinforcement-learning-based_2023}. However, these methods add computational overhead and complexity, diluting one of the original appeals of BIAs-their simplicity. Developing lightweight, problem-independent parameter control remains an open research gap.

\subsection{Limited Adaptability in Dynamic Environments}

Despite being inspired by natural systems that thrive in changing conditions, most BIAs are poorly adapted to dynamic optimization problems. Standard implementations assume static landscapes, causing swarms or populations to stagnate once optima shift \cite{jin_evolutionary_2005, mavrovouniotis_survey_2017}. Although techniques such as memory-based archives, hypermutation, multi-swarm frameworks, and prediction-based tracking have been proposed \cite{farina_dynamic_2004, li_dynamic_2025}, they remain ad hoc and lack standardization.  

Moreover, dynamic benchmarks (e.g., Moving Peaks Benchmark) only partially reflect real-world scenarios such as adaptive routing, robotics, or online portfolio selection \cite{branke_memory_1999, morrison_test_1999}. Without unified benchmarks and metrics for responsiveness and resilience, the adaptability gap will remain a barrier to broader deployment.

\subsection{Benchmarking Inconsistencies and Reproducibility}

Another systemic weakness lies in benchmarking. Much of the BIA literature relies on small sets of low-dimensional functions (Sphere, Rastrigin, Ackley), which fail to capture real-world complexity \cite{bartz-beielstein_experimental_2010}. Cross-study comparisons are further undermined by inconsistent experimental protocols, varied stopping criteria, and selective reporting of best runs rather than statistical distributions \cite{molina_comprehensive_2020, karafotias_parameter_2015}.  

The lack of rigorous, standardized evaluation has contributed to the illusion of novelty in many metaphor-driven algorithms. In reality, many claimed improvements vanish under robust statistical testing \cite{eiben_parameter_2011, hansen_coco_2021}. To restore credibility, the field must converge on community-driven benchmarks (e.g., CEC testbeds, COCO platform) and adopt reproducibility standards such as multiple independent runs, non-parametric statistical tests, and transparent reporting of full configurations.

Taken together, these challenges explain both the initial success and subsequent decline of BIAs. Their rise was fueled by creative metaphors and surprising applicability, but their long-term credibility has been undermined by poor scalability in high-dimensional problems, weak theoretical underpinnings, high sensitivity to parameter settings, limited adaptability to dynamic environments, and inconsistent benchmarking practices. To move forward, future research must shift its focus away from inventing new metaphors and instead concentrate on developing unified theoretical frameworks for convergence and parameter behavior, designing scalable algorithms that can operate effectively in dynamic and high-dimensional settings, and establishing community-driven benchmarks with reproducibility standards. Hybridization with machine learning and physics-based models also represents a promising pathway for extending capabilities in practical domains. In this sense, the study of BIAs remains valuable not as a frontier of optimization itself, but as a meta-lesson in the lifecycle of algorithmic paradigms: creative metaphors may spark innovation, but only rigorous theory and reproducible validation can sustain long-term impact.

\section{Future Directions}
\label{sec:future}

The historical trajectory of BIAs highlights both their initial creativity and their subsequent decline, offering lessons for their future development. While early algorithms such as GA, ES, DE, PSO, and ACO introduced genuine methodological novelty, the unchecked proliferation of metaphor-driven variants with little operator-level innovation diluted credibility and shifted the field away from mainstream optimization research \cite{sorensen_metaheuristicsmetaphor_2015,aranha2022metaphor,camacho2023exposing}. Thus, the future of BIAs must not lie in creating new metaphors, but rather in consolidating their theoretical foundation, improving scalability, integrating with modern AI paradigms, and restoring reproducibility through rigorous benchmarking \cite{osaba2021tutorial}.

One of the most pressing research needs is the development of stronger theoretical underpinnings. Most BIAs continue to function as heuristic black boxes, with convergence proofs that are either asymptotic or based on idealized assumptions such as infinite populations. While models based on Markov chains, dynamical systems, and Lyapunov stability have been applied to algorithms like GA and PSO \cite{poli_particle_2007,clerc_particle_2002,sudholt_new_2011}, they remain narrow in scope and fail to generalize across realistic, noisy, or constrained optimization settings. Without unified theoretical models, BIAs will remain difficult to analyze or explain, limiting their reliability in critical domains such as healthcare, robotics, and safety-critical engineering systems \cite{yang_nature-inspired_2020,molina_comprehensive_2020}. Future research must therefore emphasize rigorous convergence analysis, landscape-based characterizations, and stochastic modeling that capture the actual dynamics of these algorithms under practical conditions \cite{auger_theory_2011,bartz-beielstein_experimental_2010}.

Equally significant are the scalability and adaptability challenges faced by BIAs in high-dimensional or dynamic environments. Many swarm-based methods degrade rapidly as dimensionality increases, lacking the capacity to identify and prioritize relevant variables \cite{pham_bio-inspired_2023}. Large populations exacerbate memory and runtime costs, undermining real-time applicability in edge computing or embedded systems \cite{fu_software_2022,gill_bio-inspired_2019}. Moreover, most BIAs remain designed for static landscapes, and often fail to adapt when the objective function or constraints shift over time \cite{jin_evolutionary_2005,mavrovouniotis_survey_2017}. Promising directions include the design of dimensionality-aware search operators, lightweight implementations suitable for resource-constrained environments, and adaptive mechanisms such as memory-based models, multi-swarm systems, or prediction-driven updates that allow populations to track moving optima more effectively \cite{farina_dynamic_2004,li_dynamic_2025}.

Another long-standing limitation is parameter sensitivity. BIAs depend heavily on control parameters such as mutation rates, crossover probabilities, inertia weights, and population sizes. Unlike classical optimization methods, these parameters rarely have theoretically grounded defaults, forcing practitioners to rely on manual trial-and-error tuning that is computationally expensive and often problem-specific \cite{eiben_parameter_2011,vesterstrom_comparative_2004}. Adaptive and self-adaptive mechanisms, in which parameters co-evolve with candidate solutions, have shown promise in reducing this dependency \cite{tanabe_success-history_2013}, as have reinforcement learning and surrogate-based approaches that can dynamically adjust parameter values during the search \cite{yin_reinforcement-learning-based_2023,hutter_automated_2019}. However, these methods remain fragmented and lack standardization. Establishing reliable, transferable parameter control strategies is essential if BIAs are to achieve robustness in real-world applications \cite{molina_comprehensive_2020,aleti_systematic_2017}.

The integration of BIAs with modern artificial intelligence offers another avenue for revitalization. Rather than proposing new metaphors, hybridization with machine learning and reinforcement learning can provide a more grounded path forward. Surrogate-assisted BIAs, for instance, use neural networks to approximate fitness landscapes and reduce computational overhead, while reinforcement learning can endow agents with the capacity to adapt parameter settings and strategies dynamically \cite{darwish_bio-inspired_2018,sureka_nature_2020}. Hybrid models that combine complementary paradigms, such as GA–PSO or WOA–DE, have demonstrated improved convergence and exploration–exploitation balance in complex, multimodal tasks \cite{ajlouni_enhancing_2025,ou2023improved,nadimi2023systematic}. Such integrations shift the field from metaphor-driven novelty toward problem-driven innovation, aligning BIAs with contemporary trends in data-driven optimization.

Finally, progress in BIAs depends on restoring methodological rigor through benchmarking and reproducibility. Much of the past criticism has stemmed from inconsistent evaluation practices, limited or simplistic test functions, and the absence of statistical rigor in performance reporting \cite{molina_comprehensive_2020,karafotias_parameter_2015}. To move forward, standardized benchmarks such as those offered by the CEC competitions, the COCO platform, or the Moving Peaks Benchmark should be adopted \cite{hansen_coco_2021,branke_memory_1999}. These should be complemented by domain-specific simulators and real-world datasets from areas such as cloud computing, genomics, and network optimization \cite{calheiros2011cloudsim,NCI_TCGA,Dua:2019}. Reproducibility can further be enhanced by developing open-source repositories that consolidate implementations, metrics, and evaluation protocols, while promoting the use of statistical tests to ensure that observed improvements are not artifacts of stochastic variation \cite{bartz-beielstein_experimental_2010,auger_theory_2011}. Only through such practices can BIAs regain credibility and establish themselves as reliable tools for scientific and engineering applications.

In conclusion, the future of BIAs depends not on expanding the list of metaphor-inspired algorithms, but on learning from their rise and fall. Their continued relevance will come from addressing theoretical and scalability gaps, embedding adaptive mechanisms, hybridizing with modern AI models, and institutionalizing reproducibility. If these directions are pursued, BIAs may yet play a constructive role in specialized domains such as energy systems, bioinformatics, and autonomous robotics, while serving as a broader reminder that creativity in algorithm design must be balanced by rigor, transparency, and empirical validity to sustain long-term impact.

\section{Conclusions}
\label{sec:conclusion}

BIAs emerged as a powerful class of metaheuristics that mimic processes of evolution, swarm intelligence, foraging, and physical or social phenomena to address nonlinear, multimodal, and high-dimensional optimization problems. Foundational approaches such as GA, ES, DE, PSO, and ACO remain enduring contributions, backed by rigorous theoretical foundations, extensive benchmarking, and widespread application across domains including engineering, computer science, bioinformatics, and networking.

However, the trajectory of BIAs also illustrates the risks of unchecked proliferation. Since the early 2000s, numerous metaphor-driven algorithms, such as Firefly, Bat, Grey Wolf, Whale, Salp Swarm, and Grasshopper optimizers, have been published, yet many have been shown to repackage existing operators from GA, PSO, or DE without introducing genuine methodological novelty. Critical analyses emphasize that this trend has diluted the field’s credibility and contributed to its decline in mainstream optimization research, where better-theorized approaches such as CMA-ES, Bayesian optimization, and hybrid machine learning solvers increasingly dominate.

Despite these criticisms, the field continues to hold value, particularly through hybrid models and integration with data-driven methods. Constructive innovations such as GA–PSO hybrids, PSO combined with neural surrogates, and WOA–DE variants demonstrate that cross-paradigm integration yields tangible performance improvements in real-world applications ranging from power systems to robotics and cloud scheduling. These developments suggest that meaningful progress in BIAs arises not from superficial metaphors but from principled synthesis of established mechanisms.

Looking ahead, the sustainability of BIAs as a research direction will depend on three priorities: (i) strengthening theoretical foundations to provide convergence guarantees and interpretability in complex environments, (ii) ensuring scalability and adaptability in high-dimensional and dynamic settings, and (iii) institutionalizing standardized benchmarking protocols to restore reproducibility and comparability. Addressing these issues is essential if BIAs are to remain relevant in the era of intelligent, data-driven optimization. Ultimately, the history of BIAs offers a broader meta-lesson in algorithmic research: while creative metaphors can inspire innovation, only rigorous theory, reproducibility, and integration with modern paradigms can secure lasting impact.




\appendix

\end{document}